\documentclass[11pt]{article}

\usepackage[preprint]{acl}

\usepackage{times}
\usepackage{latexsym}

\usepackage[T1]{fontenc}

\usepackage[utf8]{inputenc}

\usepackage{microtype}

\usepackage{inconsolata}

\usepackage{times}
\usepackage{latexsym}

\usepackage[utf8]{inputenc}
\usepackage{amsmath,amssymb}
\usepackage{pifont}
\usepackage{placeins} 
\usepackage{booktabs}
\usepackage{multirow}
\usepackage{lipsum}
\usepackage{tabularx}
\usepackage{subcaption}
\usepackage{adjustbox} 
\usepackage{makecell}

\usepackage{graphicx}
\usepackage{tikz}
\usetikzlibrary{fadings} 
\usepackage{wrapfig}
\usepackage{svg}
\usepackage{multirow}

\usepackage[utf8]{inputenc}
\usepackage[table]{xcolor} 
\usepackage{array}
\usepackage[most]{tcolorbox}
\usepackage{tcolorbox}
\usepackage{tabularx}
\usepackage{adjustbox}
\usepackage{makecell}
\usepackage{inconsolata}
\usepackage{dblfloatfix}
\usepackage{enumitem}
\usepackage{cuted}
\usepackage{capt-of} 
\usepackage{dblfloatfix} 
\usepackage{caption}  
\usepackage{wrapfig}
\usepackage{fontawesome5} 
\usepackage{hyperref}     

\usepackage{graphicx}
\usepackage{booktabs}
\usepackage{multirow}
\usepackage[table]{xcolor}
\usepackage[most]{tcolorbox}
\usepackage{arydshln}
\definecolor{TableStripe}{RGB}{246,248,250}
\definecolor{TSRHighlight}{RGB}{232,244,255}
\definecolor{PromptBack}{RGB}{248,250,252}
\definecolor{PromptFrame}{RGB}{148,163,184}
\newenvironment{compactitemize}
  {\begin{itemize}
   \setlength{\itemsep}{1pt}
   \setlength{\parsep}{0pt}
   \setlength{\topsep}{2pt}
   \setlength{\partopsep}{0pt}}
  {\end{itemize}}
\newtcolorbox{promptbox}[1][]{
  enhanced,
  breakable,
  colback=PromptBack,
  colframe=PromptFrame,
  boxrule=0.5pt,
  arc=2pt,
  left=5pt,
  right=5pt,
  top=5pt,
  bottom=5pt,
  fonttitle=\bfseries\small,
  coltitle=black,
  #1
}

\newcommand{\methodname}{EvoSpatial}

\newcommand{\papertitle}{EvoSpatial: Experience-Driven Skill and Tool Evolution \\ for Spatial Reasoning Agents}

\title{\papertitle{}}

\author{
 \textbf{Bin Yu\textsuperscript{1,2,8,*}},
 \textbf{Shijie Lian\textsuperscript{2,4,8,*}},
 \textbf{Zhaolong Shen\textsuperscript{2,6,8,*}},
 \textbf{Xiaopeng Lin\textsuperscript{7,8,*}},
 \textbf{Yao Zhang\textsuperscript{2,5}},
\\
 \textbf{Haishan Liu\textsuperscript{2}},
 \textbf{Haotian Li\textsuperscript{1}},
 \textbf{Yuliang Wei\textsuperscript{1,$\dagger$}},
 \textbf{Kai Hu\textsuperscript{8}},
 \textbf{Yichao Du\textsuperscript{8}},
\\
 \textbf{Zhirui Zhang\textsuperscript{8}},
 \textbf{Bailing Wang\textsuperscript{1}},
 \textbf{Kai Chen\textsuperscript{2,3,8,$\dagger$}}
\\
 {\small \textsuperscript{1}Harbin Institute of Technology
 \textsuperscript{2}Zhongguancun Academy}
 {\small \textsuperscript{3}Zhongguancun Institute of Artificial Intelligence}
\\
 {\small \textsuperscript{4}Huazhong University of Science and Technology}
 {\small \textsuperscript{5}Wuhan University}
 {\small \textsuperscript{6}Beihang University}
\\
 {\small \textsuperscript{7}The Hong Kong University of Science and Technology (Guangzhou)}
 {\small \textsuperscript{8}DeepCybo}
}

\begin{document}
\maketitle
\makeatletter
\begingroup
\renewcommand{\thefootnote}{\fnsymbol{footnote}}
\renewcommand{\@makefntext}[1]{\noindent\makebox[1em][l]{\@makefnmark}#1}
\footnotetext[1]{Equal contribution.}
\footnotetext[2]{Corresponding authors.}
\endgroup
\begingroup
\renewcommand{\thefootnote}{}
\renewcommand{\@makefntext}[1]{\noindent#1}
\footnotetext{This work was conducted during an internship at DeepCybo.}
\endgroup
\makeatother
\setcounter{footnote}{0}

\begin{abstract}
Spatial reasoning agents can inspect tool outputs and repair errors, but reusing these lessons requires preserving both the computation and the conditions that make it valid. We introduce \textbf{\methodname{}}, a framework that learns from execution experience by updating a library of linked skills and tools. Skills describe procedures, geometric assumptions, and verification steps; tools implement the operations those procedures require. When solving a task, the agent retrieves relevant skills and synthesizes reusable computational functions if existing tools are insufficient. After each task, execution feedback and evidence checks guide skill revision and tool consolidation, without ground-truth answers or benchmark scores. Across 20 benchmarks, \methodname{} averages 63.9, 4.0 percentage points above the average reported by SpatialClaw. The evolved library also improves held-out and cross-benchmark performance when updates are disabled, and trajectories generated with evolved skills improve supervised fine-tuning, even without external skills at evaluation. Keeping checked procedures on record, together with the tools that support them, lets an agent carry spatial reasoning experience from one task to the next.
\end{abstract}

\section{Introduction}
\label{sec:intro}

\makeatletter
\if@twocolumn
    \begin{figure}[!t]
\else
    \begin{wrapfigure}{r}{0.38\textwidth}
\fi
    \centering
    \includegraphics[width=\linewidth]{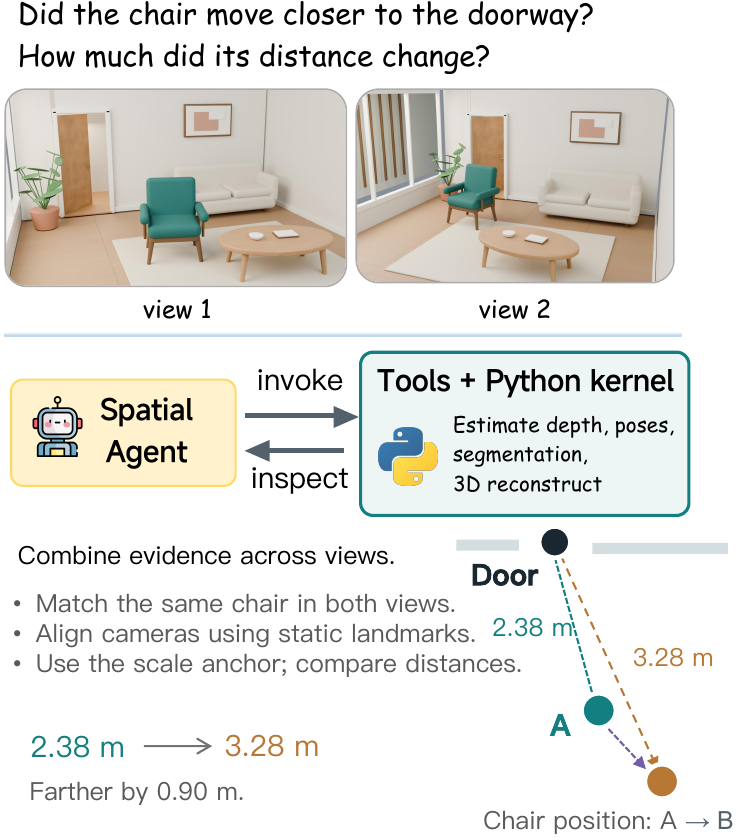}
\if@twocolumn
    \caption{\textbf{Interactive spatial reasoning across views.}
    Tools support correspondence, frame alignment, and distance comparison.}
\else
    \caption{\textbf{Interactive spatial reasoning across views.}
    Synthetic scene geometry with a known 1.10\,m threshold width.}
\fi
    \label{fig:spatial-agent-example}
\if@twocolumn
    \end{figure}
\else
    \end{wrapfigure}
\fi
\makeatother

Did a chair move toward a doorway, or did the camera simply move? Answering this question across two views requires separating object motion from camera motion, aligning reference frames, and comparing distances (Figure~\ref{fig:spatial-agent-example}). Spatial agents augment vision-language models (VLMs) with segmentation, depth estimation, and reconstruction tools to obtain the required geometric evidence~\cite{chen2026spacetools,luo2026pyspatial,cho2026spatialclaw}. The agent must then choose compatible operations, check their assumptions, and determine whether the measurements support an answer.

Existing spatial agents learn tool coordination or compose operations through visual programs and interactive execution~\cite{chen2026spacetools,luo2026pyspatial,cho2026spatialclaw}. Interactive execution allows an agent to inspect intermediate results and repair errors before answering. A repaired error can also provide a lesson for later tasks: which computation to use, which assumption failed, or which check prevented a mistake. We ask how a spatial reasoning agent can retain and apply these lessons across tasks.

Self-evolving agents refine their procedures and executable capabilities from interaction experience~\cite{li2026yunjue,ouyang2025reasoningbank,zhang2026coevoskills}. In spatial reasoning, these procedures must account for the conditions under which geometric measurements are valid. A coordinate conversion must match the reference frames being compared; separated object centers do not rule out contact between their surfaces. Reusable spatial experience therefore needs to record both the computation and the assumptions that justify its use. It must also specify how to check those assumptions on a new task.

\begin{figure}[!ht]
    \centering
    \includegraphics[width=\linewidth]{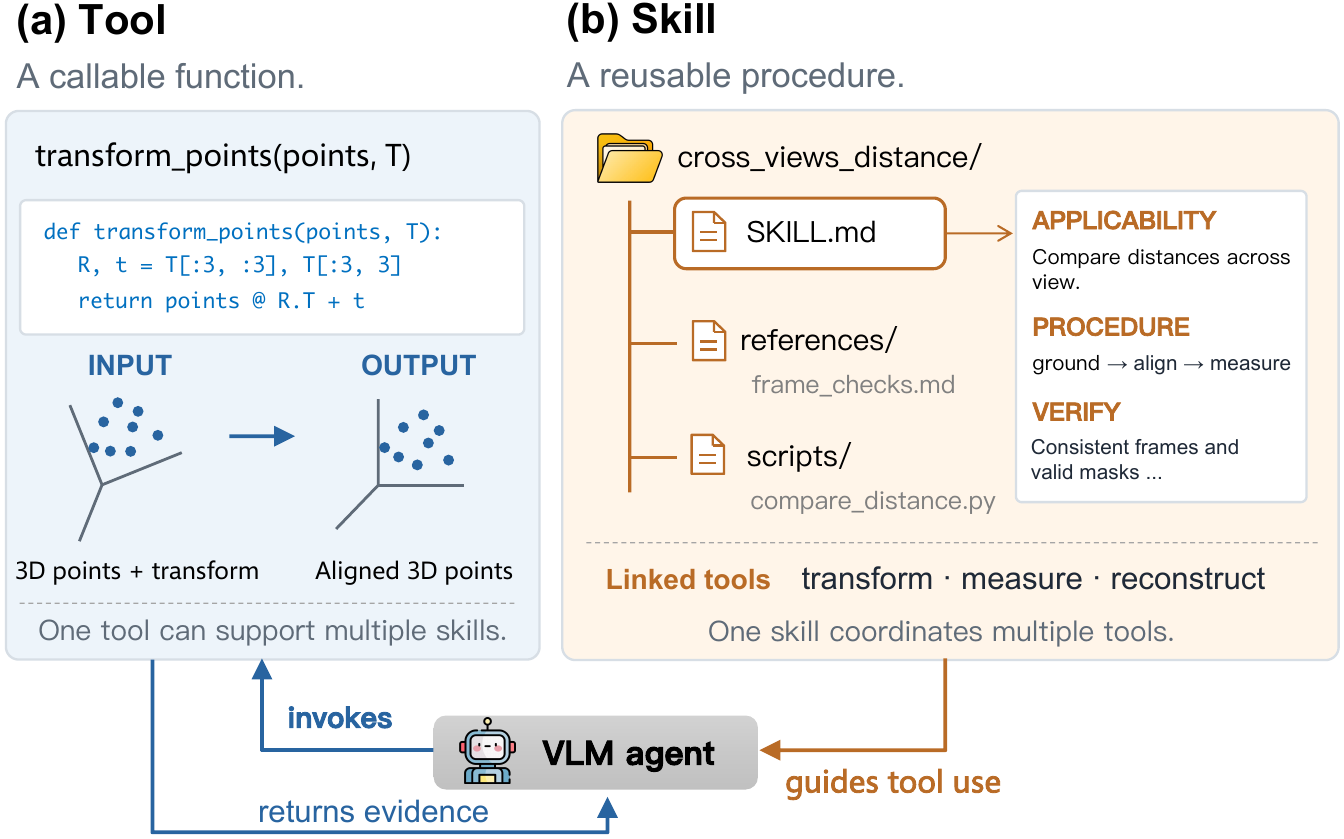}
    \caption{\textbf{Tools and skills in \methodname{}.}
    Tools provide callable operations; skills specify how to combine them and check the resulting evidence.}
    \label{fig:skill-tool}
\end{figure}

\begin{figure*}[t]
    \centering
    \includegraphics[width=\textwidth]{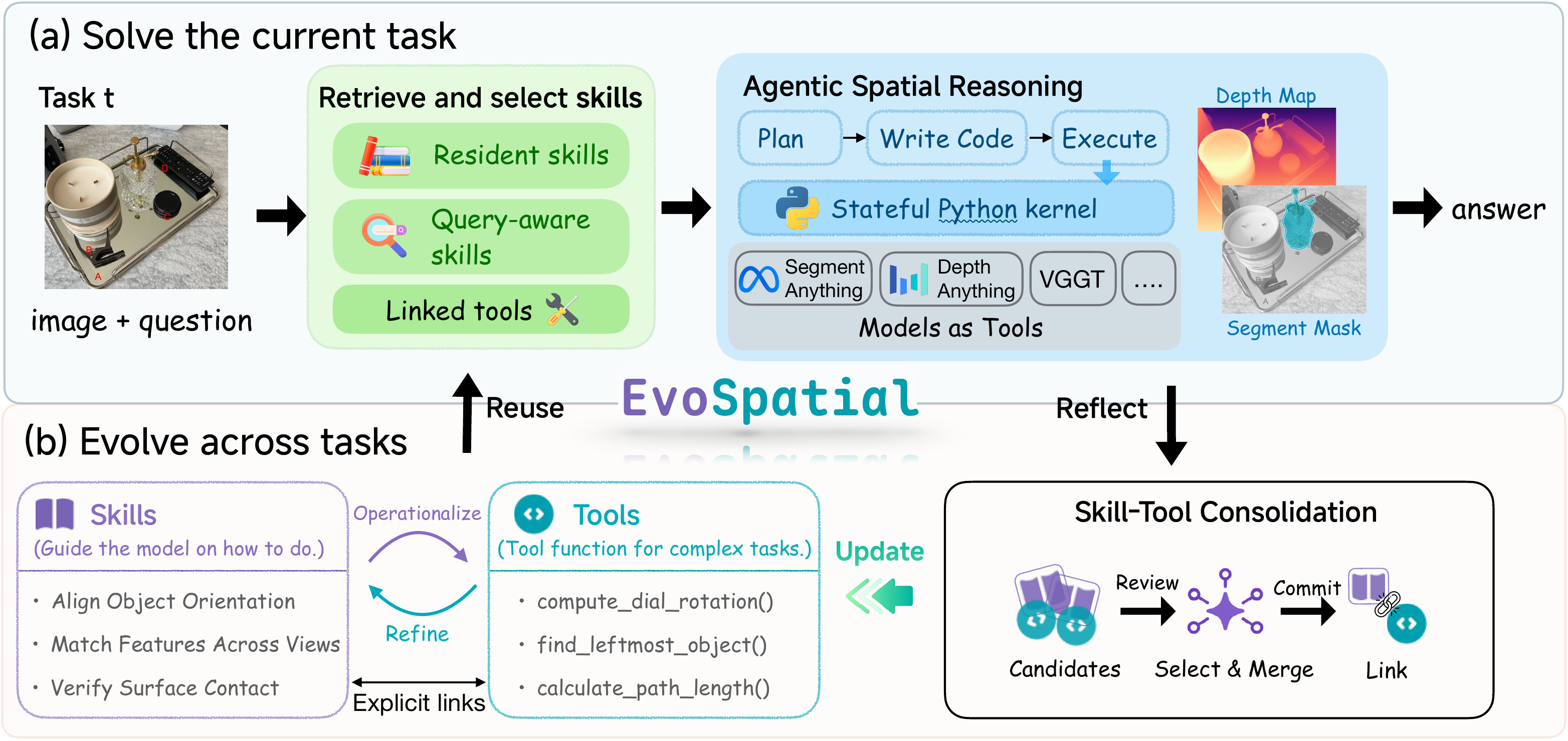}
    \caption{\textbf{Overview of \methodname{}.}
    (a) Resident and retrieved skills guide interactive spatial reasoning with linked tools.
    (b) Execution feedback guides helper synthesis, reflection, and library updates for subsequent tasks.}
    \label{fig:evospatial-overview}
\end{figure*}

We introduce \textbf{\methodname{}}, a framework that updates spatial reasoning skills and tools from execution experience (Figure~\ref{fig:evospatial-overview}). A \emph{tool} exposes a callable operation, either through a perception model or a computational helper. We call a reusable function synthesized by the agent a \emph{generated helper}. A \emph{skill} describes a multi-step procedure, its conditions of use, and the evidence required to verify its outcome (Figure~\ref{fig:skill-tool}). A persistent \emph{skill--tool library} links each skill to the tools it uses.

For each task, \methodname{} selects relevant skills and generates task-specific usage guidance. When existing tools cannot perform a required computation, it synthesizes a helper. The agent executes code, inspects the results, and repairs errors before submitting an answer. Afterwards, it extracts reusable lessons from checked operations and observed repairs, revises skills, and consolidates generated helpers; the updated library is then available to subsequent tasks.

We evaluate \methodname{} on 20 benchmarks: it reaches an average score of 63.9, 4.0 percentage points above SpatialClaw's reported average. To separate reuse from continued adaptation, we also evaluate with the evolved library fixed. Jointly evolved skills and tools raise the held-out average from 62.0 to 65.4. Libraries evolved on ERQA and MindCube improve the average on 18 common target benchmarks by 1.73 and 1.76 points over the initial library, respectively. Finally, we test whether trajectories generated with evolved skills improve supervised fine-tuning without external skills at evaluation.

We make three contributions:
\begin{enumerate}[leftmargin=1.3em,itemsep=0pt,parsep=0pt,partopsep=0pt,topsep=2pt]
    \item We introduce \textbf{\methodname{}}, which retains spatial reasoning experience as linked skills and tools, together with geometric assumptions and verification steps.
    \item We combine task-specific usage guidance and helper synthesis with post-task reflection and library curation; the evolved library then improves held-out and cross-benchmark performance even with updates disabled at evaluation.
    \item We report a 4.0-point average gain over SpatialClaw across 20 benchmarks. Further experiments assess fixed-library generalization, cross-benchmark transfer, individual components, and learning from skill-guided trajectories.
\end{enumerate}

\section{Related Work}
\label{sec:related}

\textbf{Agent skills.}
Agent skills package procedures and supporting resources for use at inference time~\cite{anthropic2025agentskills,xu2026agentskills,lu2026skillzero}. Prior work acquires skills from interaction and execution feedback~\cite{zheng2025skillweaver,ouyang2025reasoningbank,zhang2025ace,zhang2026coevoskills}, then revises and maintains them across tasks~\cite{wang2026sage,xia2026skillrl,li2026codeskill}. Related approaches build executable capability libraries and organize skills for retrieval and composition~\cite{li2026yunjue,li2026agentskillos,yang2026skillseek}. Recent evaluations test skill utility, robustness, and transfer across tasks and agent configurations~\cite{li2026skillsbench,jiang2026demystifying,piao2026heldout,liu2026skillscriptbench}. These advances have been studied primarily in coding, web navigation, and everyday digital tasks~\cite{li2026codeskill,zheng2025skillweaver,wang2026sage}. Their use in visually grounded 3D spatial reasoning, a capability central to embodied intelligence, remains comparatively less explored. We extend skill evolution to this setting, where reusable procedures must account for the assumptions behind perception outputs and geometric computations.

\textbf{Tool-augmented spatial reasoning.}
Specialist perception models~\citep{ME-VLM,physbrain1.5,physbrain1.0,RoboBrain2.5,EuclidGift} provide geometric representations that complement direct VLM predictions. SpaceTools learns spatial tool coordination through interactive training~\cite{chen2026spacetools}, while pySpatial composes spatial operations in visual programs~\cite{luo2026pyspatial}. SpatialClaw uses multi-turn execution to invoke tools, inspect their outputs, and adjust subsequent actions~\cite{cho2026spatialclaw}. \methodname{} builds on interactive execution by retaining its lessons in a skill--tool library that changes across tasks.

\section{Method}
\label{sec:method}

\methodname{} alternates between solving a task and updating its persistent skill--tool library $\mathcal{L}_t$ (Figure~\ref{fig:evospatial-overview}). Given a question $q_t$ and visual inputs $I_t$ (images or sampled video frames), the agent returns an answer $\hat y_t$ and an execution trajectory $\tau_t$. It then updates the library using execution feedback $f_t$:
\begin{equation}
    \begin{aligned}
    (\hat y_t,\tau_t) &= \operatorname{Agent}(q_t,I_t,\mathcal{L}_t),\\
    \mathcal{L}_{t+1} &= \operatorname{Update}(\mathcal{L}_t,\tau_t,f_t).
    \end{aligned}
    \label{eq:evolution-loop}
\end{equation}
The trajectory $\tau_t$ records executed code, observations, repairs, and the submitted answer. Feedback $f_t$ comes from this record and checks of the visual and geometric evidence. Library updates use observed capability gaps, execution failures, and locally verified improvements; they receive neither ground-truth answers nor benchmark correctness scores.

\subsection{Interactive Spatial Reasoning}
\label{sec:execution}

\paragraph{Perception tools.}
Perception tools provide object regions through grounding and segmentation, and geometric representations through depth estimation and reconstruction~\cite{chen2026spacetools,luo2026pyspatial}. These representations include depth maps, camera parameters, and 3D point maps, from which the agent computes distances and other spatial quantities. Table~\ref{tab:perception-tools} lists the underlying models and their capabilities.

\begin{table}[!ht]
\centering
\footnotesize
\setlength{\tabcolsep}{3pt}
\renewcommand{\arraystretch}{1.12}
\begin{tabular}{@{}>{\centering\arraybackslash}m{0.25\columnwidth}>{\raggedright\arraybackslash}m{0.43\columnwidth}>{\raggedright\arraybackslash}m{0.26\columnwidth}@{}}
\toprule
Model & Capability & Reference \\
\midrule
\rowcolor{black!6}
Depth Anything 3 & Depth, camera parameters, and 3D reconstruction & \citep{depthanything3} \\
VGGT & Camera parameters, depth maps, and 3D point maps & \citep{VGGT_25} \\
\rowcolor{black!6}
Pi3X & Multi-view 3D reconstruction & \citep{wang2025pi} \\
MapAnything & Metric 3D reconstruction & \citep{keetha2026mapanything} \\
\rowcolor{black!6}
SAM 3 & Image segmentation and video segmentation/tracking & \citep{carion2025sam} \\
SAM 3.1 Object Multiplex & Multi-object video segmentation/tracking & \citep{carion2025sam} \\
\bottomrule
\end{tabular}
\normalsize
\caption{\textbf{Perception models available to the agent.}}
\label{tab:perception-tools}
\end{table}

\paragraph{Stateful execution.}
The VLM executes code in a task-local Python kernel that retains intermediate outputs across steps~\cite{cho2026spatialclaw}. To compare distances across views, for example, it can extract reconstructed points using object masks, transform them into a common reference frame, and compute distances. The agent chooses the operations and their order for each task.

\paragraph{Plan, execute, and inspect.}
A planner identifies the evidence needed to answer the question, given the available tools and selected skills. The execution agent then alternates between writing code and inspecting its outputs. Measurements, error messages, and visualizations such as mask overlays enter the VLM's context; dense intermediate arrays remain in the kernel. The agent uses this feedback to revise its grounding or computation before answering. Kernel state is discarded after each task, while skills and generated helpers persist in the library.

\subsection{Skill--Tool Experience Reuse}
\label{sec:library}

The library links skills to perception tools, initial computational helpers, and generated helpers through documented interfaces. A helper can support multiple skills, and a skill can use several tools. Records of prior use and assessed utility inform both retrieval and curation.

The \textbf{Skill Selector} retrieves candidates through lexical and semantic matching, then selects complementary skills using task requirements and recorded reliability. For each selected skill, it generates \emph{task-specific usage guidance} identifying applicable steps, unresolved requirements, and evidence to check before answering. The agent receives the selected skills, guidance, and linked tools, together with a small resident set of trusted skills. The selector may return no candidates when none is suitable.

The \textbf{Tool Synthesizer} checks whether existing tools, alone or in combination, can perform the required computation. If they cannot, it synthesizes a generated helper, following demand-driven tool evolution~\cite{li2026yunjue}. For example, a helper for cross-view measurement might transform reconstructed points into the reference frame required by a skill.

During execution, recurring errors prompt retrieval of repair guidance, while missing operations trigger helper synthesis or refinement. Helpers that pass admission checks become available to the current task and are considered for later reuse (Section~\ref{sec:evolution}). Appendix~\ref{app:implementation} gives the retrieval settings, admission checks, and generation budgets.

\subsection{Experience Reflection and Evolution}
\label{sec:evolution}

After each task, the \textbf{Experience Reflector} extracts reusable procedures from execution feedback~\cite{li2026yunjue,ouyang2025reasoningbank}. It reviews checked operations, errors revealed by execution or conflicting evidence, and repairs that resolved those errors. For example, an empty text-prompted mask may be repaired through box-guided segmentation. Once the resulting region has been visually checked, this repair can become a fallback procedure. Each candidate records its conditions of use, verification steps, and supporting observations.

The reflector distinguishes supported observations from contradicted assumptions and unresolved judgments. A tool call that completes without error does not by itself establish a correct spatial inference. Relevant checks include compatible coordinate frames, valid measurement scales, and agreement between a mask and the visible object. When evidence is inconclusive, the reflector retains only supported local lessons or proposes no update. Figure~\ref{fig:evolution-case-study} gives a contact-verification example: separated centroids do not rule out contact, so the agent must inspect local surfaces and gaps.

The \textbf{Library Curator} incorporates candidate skills and generated helpers into the library. Following incremental context adaptation~\cite{zhang2025ace}, it makes local skill edits and preserves existing conditions and checks unless evidence supports a change. It also uses function signatures and comments to identify overlapping operations and merge related implementations, following interface-guided tool consolidation~\cite{li2026yunjue}. New and consolidated helpers undergo execution checks on representative inputs before retention.

Reliability is assessed separately for each observed application of a skill or helper. An LLM judge records support, contradiction, or insufficient evidence for the intended use, linking its assessment to logged calls, outputs, and checks. New entries remain provisional until repeated supported applications justify wider reuse. Contradictory evidence can trigger revision or suspension; inconclusive applications leave reliability unchanged. These records guide selection and curation even when the final answer cannot be verified. Appendix~\ref{app:implementation} specifies the assessment and library-management rules.

\section{Experiments}
\label{sec:experiments}

We evaluate whether accumulated experience improves spatial reasoning beyond access to perception tools and interactive execution. The experiments compare task-solving performance, test reuse with a fixed library, and isolate the roles of usage guidance and curation. We also examine cross-benchmark transfer and training on skill-guided trajectories.

\begin{table*}[t]
\centering
\footnotesize
\setlength{\tabcolsep}{3pt}
\renewcommand{\arraystretch}{1.06}
\begin{adjustbox}{max width=\textwidth}
\begin{tabular}{@{}lcccccccc@{}}
\toprule
\makecell[cl]{Benchmark} & \makecell[c]{No-tool} & \makecell[c]{Single-Pass\\Code} & \makecell[c]{Structured\\Tool-Call} & \makecell[c]{VADAR\\\citeyearpar{marsili2025visual}} & \makecell[c]{pySpatial\\\citeyearpar{luo2026pyspatial}} & \makecell[c]{SpaceTools\\\citeyearpar{chen2026spacetools}} & \makecell[c]{SpatialClaw\\\citeyearpar{cho2026spatialclaw}} & \makecell[c]{\textbf{\methodname{}}\\(Ours)} \\
\midrule
\multicolumn{9}{l}{\cellcolor{TSRHighlight}\textit{Single-image spatial reasoning}} \\
ERQA~\citep{team2025gemini} & 58.3 & 58.3 & 59.0 & 33.3 & 50.8 & 52.0 & 61.3 & 63.5$_{\pm 0.25}$ \\
Omni3D~\citep{marsili2025visual} & 51.7 & 48.2 & 55.7 & 44.6 & 50.6 & 50.7 & 54.3 & 58.1$_{\pm 0.42}$ \\
OmniSpatial~\citep{jia2025omnispatial} & 57.3 & 60.0 & 59.6 & 42.4 & 58.0 & 60.4 & 63.6 & 67.0$_{\pm 0.36}$ \\
SPBench~\citep{li2025spatialladder} & 55.1 & 58.2 & 60.5 & 41.6 & 53.5 & 45.1 & 68.4 & 71.0$_{\pm 0.40}$ \\
\addlinespace[3pt]
\multicolumn{9}{l}{\cellcolor{TSRHighlight}\textit{Multi-view spatial reasoning}} \\
MindCube~\citep{yin2025mindcube} & 57.5 & 57.5 & 62.4 & \textemdash{}$^{*}$ & 67.1 & 52.9 & 72.8 & 78.4$_{\pm 0.72}$ \\
MMSI~\citep{yang2025mmsi} & 37.9 & 42.3 & 43.0 & \textemdash{}$^{*}$ & 33.2 & 33.1 & 51.3 & 54.0$_{\pm 0.61}$ \\
SPAR-Bench~\citep{zhang2025from} & 55.2 & 61.1 & 58.7 & \textemdash{}$^{*}$ & 51.7 & 53.9 & 63.3 & 68.8$_{\pm 0.58}$ \\
\addlinespace[3pt]
\multicolumn{9}{l}{\cellcolor{TSRHighlight}\textit{Video spatial \& 4D reasoning}} \\
MMSI-Video~\citep{lin2025mmsivideo} & 36.9 & 37.1 & 35.1 & \textemdash{}$^{*}$ & 28.7 & 36.6 & 41.6 & 48.4$_{\pm 0.70}$ \\
OSI-Bench~\citep{wu2025osibench} & 35.6 & 36.8 & 40.3 & \textemdash{}$^{*}$ & 32.0 & 30.2 & 41.9 & 48.2$_{\pm 0.67}$ \\
PAI-Bench~\citep{zhou2025paibench} & 65.0 & 64.2 & 65.6 & \textemdash{}$^{*}$ & 46.0 & 65.1 & 68.1 & 72.0$_{\pm 0.44}$ \\
VSI-Bench-U~\citep{brown2025shortcuts} & 48.0 & 47.8 & 50.1 & \textemdash{}$^{*}$ & 44.4 & 33.6 & 48.5 & 54.2$_{\pm 0.71}$ \\
VSTI-Bench~\citep{fan2025vlm} & 54.7 & 64.2 & 63.5 & \textemdash{}$^{*}$ & 55.0 & 56.0 & 67.6 & 72.7$_{\pm 0.64}$ \\
DSI-Bench~\citep{zhang2025dsibench} & 45.3 & 57.9 & 58.4 & \textemdash{}$^{*}$ & 43.7 & 43.0 & 62.9 & 69.0$_{\pm 0.53}$ \\
\addlinespace[3pt]
\multicolumn{9}{l}{\cellcolor{TSRHighlight}\textit{General spatial reasoning}} \\
BLINK~\citep{fu2024blink} & 75.7 & 73.9 & 73.9 & \textemdash{}$^{*}$ & 60.3 & 58.2 & 73.4 & 74.8$_{\pm 0.27}$ \\
SpatialTree~\citep{xiao2025spatialtree} & 59.9 & 58.9 & 57.7 & \textemdash{}$^{*}$ & 53.9 & 52.7 & 60.7 & 64.3$_{\pm 0.37}$ \\
ViewSpatial~\citep{li2025viewspatial} & 51.7 & 52.0 & 55.5 & \textemdash{}$^{*}$ & 49.9 & 52.1 & 60.2 & 64.4$_{\pm 0.44}$ \\
\addlinespace[3pt]
\multicolumn{9}{l}{\cellcolor{TSRHighlight}\textit{General video understanding}} \\
CVBench~\citep{zhu2025cvbench} & 69.8 & 71.3 & 73.6 & \textemdash{}$^{*}$ & 67.7 & 46.8 & 72.2 & 74.0$_{\pm 0.17}$ \\
PerceptionComp~\citep{li2026perceptioncomp} & 36.8 & 39.2 & 42.5 & \textemdash{}$^{*}$ & 37.1 & 38.6 & 44.0 & 48.1$_{\pm 0.61}$ \\
Video-MME~\citep{fu2025video} & 74.8 & 74.6 & 75.8 & \textemdash{}$^{*}$ & 50.3 & 74.6 & 77.0 & 78.2$_{\pm 0.27}$ \\
Video-MME-v2~\citep{videommev2_2026} & 40.7 & 40.2 & 44.0 & \textemdash{}$^{*}$ & 23.0 & 37.4 & 44.4 & 48.9$_{\pm 0.53}$ \\
\midrule
Average (20 benchmarks) & 53.4 & 55.2 & 56.7 & \textemdash{}$^{*}$ & 47.8 & 48.7 & 59.9 & \textbf{63.9} \\
\bottomrule
\end{tabular}
\end{adjustbox}
\normalsize
\caption{\textbf{Performance across 20 benchmarks with Gemma4-31B.}
Baseline scores are reported by SpatialClaw~\cite{cho2026spatialclaw}. Subscripts denote sample standard deviations over three runs of \methodname{}; $^{*}$ marks unsupported video or multi-image inputs.}
\label{tab:spatial-main-results}
\end{table*}

\subsection{Experimental Setup}
\label{sec:experimental-setup}

\textbf{Benchmarks and metrics.}
We evaluate 20 benchmarks across five task groups (Table~\ref{tab:spatial-main-results}), using all examples when a benchmark contains at most 1,000 and a fixed-seed sample of 1,000 otherwise. Categorical questions use accuracy; numerical questions retain benchmark-specific metrics (Appendix~\ref{app:benchmark-scoring}). Scores are percentages averaged over examples within each benchmark. Category and overall scores equally weight the corresponding benchmarks.

\textbf{Backbone and execution.}
The main comparison uses Gemma4-31B~\cite{gemma4}. We preserve image aspect ratios and limit the longer edge to 768 pixels. The initial visual context contains at most 32 frames, sampled uniformly for longer videos; multi-image inputs retain the first 32 images. Interactive agents may take up to 30 tool-call turns per task.

\textbf{Comparison methods.}
\emph{No-tool} answers directly from visual inputs. \emph{Single-Pass Code}, \emph{VADAR}~\cite{marsili2025visual}, and \emph{pySpatial}~\cite{luo2026pyspatial} generate programs without intermediate feedback. \emph{Structured Tool-Call} and \emph{SpaceTools Toolshed}~\cite{chen2026spacetools} interact through named JSON tool calls, whereas \emph{SpatialClaw}~\cite{cho2026spatialclaw} executes code iteratively in a persistent kernel. VADAR supports only single-image inputs.

\textbf{Repeated runs.}
We conduct three independent runs. In the main comparison, \methodname{} evolves online from execution feedback, starting with a fresh initial library for each benchmark and run. The runs vary the order of questions; Appendix~\ref{app:main-evolution-protocol} specifies the feedback and reset protocol. We report means and sample standard deviations for \methodname{} in the main comparison, and means for held-out, transfer, and ablation results. Category and overall averages are computed within each run before aggregation across runs.

\subsection{Main Results}
\label{sec:main-comparison}

Table~\ref{tab:spatial-main-results} compares baseline scores reported by SpatialClaw~\cite{cho2026spatialclaw} with our three-run results. \methodname{} reaches the best overall average (63.9), $4.0$ percentage points above SpatialClaw. The largest category gains appear in video spatial and 4D reasoning ($+5.7$ points) and in multi-view reasoning ($+4.6$). MMSI-Video and MindCube improve by $6.8$ and $5.6$ points, respectively. These tasks require combining evidence across views or time, the setting where reusable spatial procedures should help most. Many BLINK questions can be answered through rapid, intuitive visual judgments without lengthy textual reasoning~\cite{fu2024blink}. All evaluated tool-augmented methods trail the no-tool baseline (75.7), but \methodname{} scores highest among them (74.8) and narrows the gap to 0.9 points, versus 2.3 for SpatialClaw. This pattern is consistent with findings that extended multimodal reasoning can weaken visual grounding and amplify hallucination~\cite{liu2025morethinking}; additional reasoning is not always beneficial for such intuitive judgments.

\subsection{Generalization to Unseen Tasks}
\label{sec:held-out-generalization}

We next test whether the library remains useful after evolution stops. This experiment uses Gemma4-31B on the 10 non-video benchmarks in Table~\ref{tab:spatial-main-results}: four single-image, three multi-view, and three general spatial benchmarks. We evolve the library on approximately 20\% of each benchmark and evaluate on the remaining 80\%. Questions sharing visual inputs stay in the same split. We compare the initial library with skill-only, tool-only, and joint evolution; each single-component variant keeps the other component at initialization. Evaluation retains retrieval and interactive execution under identical budgets, but disables library updates and helper synthesis. All variants share the partition and initial library across three independent runs.

Joint evolution raises the held-out average from 62.0 to 65.4, compared with 63.8 for skill-only and 64.0 for tool-only evolution (Table~\ref{tab:held-out-generalization}). It performs best in all three task groups, with the largest gain over the initial library in multi-view reasoning ($+4.1$ points). Since neither single-component variant reaches this level, skills and generated helpers appear to contribute complementary benefits on unseen tasks.

\begin{table}[!ht]
\centering
\footnotesize
\setlength{\tabcolsep}{2.5pt}
\renewcommand{\arraystretch}{1.12}
\begin{adjustbox}{max width=\columnwidth}
\begin{tabular}{@{}lcccc@{}}
\toprule
Library & \makecell{Single-\\image} & \makecell{Multi-\\view} & General & \makecell{Avg.\\(10)} \\
\midrule
Initial & 61.0 & 61.3 & 64.1 & 62.0 \\
\rowcolor{black!6}
Evolved skills only & 62.6 & 63.4 & 65.9 & 63.8 \\
Evolved tools only & 63.1 & 63.7 & 65.5 & 64.0 \\
\rowcolor{black!6}
\textbf{Evolved skills \& tools} & 64.2 & 65.4 & 66.9 & 65.4 \\
\bottomrule
\end{tabular}
\end{adjustbox}
\caption{\textbf{Generalization to unseen tasks.} The library evolves on approximately 20\% of each benchmark and remains fixed during evaluation on the remaining 80\%.}
\label{tab:held-out-generalization}
\end{table}

\begin{table*}[t]
\centering
\footnotesize
\setlength{\tabcolsep}{2.5pt}
\renewcommand{\arraystretch}{1.15}
\begin{adjustbox}{max width=\textwidth}
\begin{tabular}{@{}ll*{11}{c}@{}}
\toprule
 & & \multicolumn{4}{c}{Single-image} & \multicolumn{3}{c}{Multi-view} & \multicolumn{3}{c}{General} & \\
\cmidrule(lr){3-6}\cmidrule(lr){7-9}\cmidrule(lr){10-12}
Base model & Method & ERQA & Omni3D & \makecell{Omni\\Spatial} & SPBench & \makecell{Mind\\Cube} & MMSI & \makecell{SPAR-\\Bench} & BLINK & \makecell{Spatial\\Tree} & \makecell{View\\Spatial} & \makecell{Avg.\\(10)} \\
\midrule
 & No-tool & 58.3 & 51.7 & 57.3 & 55.1 & 57.5 & 37.9 & 55.2 & 75.7 & 59.9 & 51.7 & 56.0 \\
\multirow{-2}{*}{Gemma4-31B} & \textbf{\methodname{}} & 63.5 & 58.1 & 67.0 & 71.0 & 78.4 & 54.0 & 68.8 & 74.8 & 64.3 & 64.4 & 66.4 \\
\addlinespace[2pt]
\rowcolor{black!6}
 & No-tool & 56.0 & 44.8 & 56.9 & 45.1 & 47.8 & 31.5 & 48.3 & 70.2 & 53.7 & 53.1 & 50.7 \\
\rowcolor{black!6}
\multirow{-2}{*}{Gemma4-26B-A4B} & \textbf{\methodname{}} & 58.5 & 52.1 & 60.4 & 65.0 & 68.7 & 45.0 & 67.0 & 72.6 & 61.0 & 62.0 & 61.2 \\
\addlinespace[2pt]
 & No-tool & 57.2 & 51.2 & 57.0 & 52.4 & 45.4 & 38.0 & 47.8 & 70.6 & 58.3 & 52.0 & 53.0 \\
\multirow{-2}{*}{Qwen3.6-35B-A3B} & \textbf{\methodname{}} & 60.5 & 57.1 & 63.0 & 66.0 & 66.7 & 48.0 & 62.5 & 75.0 & 63.3 & 56.5 & 61.9 \\
\addlinespace[2pt]
\rowcolor{black!6}
 & No-tool & 59.5 & 55.9 & 60.8 & 58.0 & 51.2 & 40.9 & 55.0 & 70.5 & 62.6 & 53.9 & 56.8 \\
\rowcolor{black!6}
\multirow{-2}{*}{Qwen3.6-27B} & \textbf{\methodname{}} & 63.5 & 60.7 & 65.9 & 67.6 & 75.0 & 55.5 & 72.3 & 78.2 & 67.5 & 62.5 & 66.9 \\
\addlinespace[2pt]
 & No-tool & 60.5 & 56.7 & 61.7 & 58.9 & 52.6 & 42.1 & 55.8 & 71.4 & 63.5 & 54.8 & 57.8 \\
\multirow{-2}{*}{Qwen3.8-27B} & \textbf{\methodname{}} & 64.5 & 61.3 & 66.7 & 68.4 & 75.8 & 56.1 & 72.8 & 78.5 & 68.2 & 63.3 & 67.6 \\
\bottomrule
\end{tabular}
\end{adjustbox}
\caption{\textbf{\methodname{} across base models.} No-tool and \methodname{} scores on 10 benchmarks.}
\label{tab:backbone-results}
\end{table*}

\subsection{Ablation Studies}
\label{sec:ablation-studies}

We examine task-specific usage guidance, geometric verification guidance, and library curation under the held-out protocol in Section~\ref{sec:held-out-generalization}. All variants use Gemma4-31B with matched data partitions and execution budgets, and keep the library fixed during evaluation. Table~\ref{tab:ablation-results} reports means over three independent runs.

\textbf{Task-specific usage guidance.}
Removing the Skill Selector's task-specific usage guidance reduces the average score from 65.4 to 64.1, with the library, selected skills, and linked tools unchanged. Scores decline in all three task groups, most sharply in multi-view reasoning ($2.0$ points). The retrieved procedure alone is less useful than the same procedure with guidance on which steps apply and what evidence to check.

\textbf{Geometric verification guidance.}
We remove geometric verification guidance from skill text and task-specific usage guidance using a fixed LLM filter. The evolved library, selected skill IDs, geometric assumptions, core computations, and tool implementations stay unchanged. The agent may still verify evidence on its own. The average score falls from 65.4 to 63.6 ($-1.8$ points), with the largest decline in multi-view reasoning ($-2.7$), compared with single-image ($-1.3$) and general reasoning ($-1.5$). This pattern supports the value of explicit verification guidance when combining evidence across views, where reference-frame, scale, and cross-view consistency checks help ensure that the retained geometric assumptions hold.

\textbf{Library curation.}
We replace the Library Curator's skill updates and tool merging with append-only storage of new skills and generated helpers. This variant uses the same initial library and adaptation data, retaining reflection, helper synthesis, and usage-based qualification. Its average score falls to 63.7 ($-1.7$ points), with declines in every task group. Revising and consolidating entries works better than simply accumulating them.

\begin{table}[!ht]
\centering
\footnotesize
\setlength{\tabcolsep}{2.5pt}
\renewcommand{\arraystretch}{1.12}
\begin{adjustbox}{max width=\columnwidth}
\begin{tabular}{@{}>{\raggedright\arraybackslash}m{0.44\columnwidth}cccc@{}}
\toprule
Ablation setting & \makecell{Single-\\image} & \makecell{Multi-\\view} & General & \makecell{Avg.\\(10)} \\
\midrule
\textbf{\methodname{} (full)} & \textbf{64.2} & \textbf{65.4} & \textbf{66.9} & \textbf{65.4} \\
\rowcolor{black!6}
Skill Selector\newline (w/o task-specific\newline usage guidance) & 63.1 & 63.4 & 66.0 & 64.1 \\
\rowcolor{white}
Geometric verification\newline (w/o verification\newline guidance) & 62.9 & 62.7 & 65.4 & 63.6 \\
\rowcolor{black!6}
Library Curator\newline (append-only) & 62.6 & 63.1 & 65.8 & 63.7 \\
\bottomrule
\end{tabular}
\end{adjustbox}
\caption{\textbf{Ablations of guidance and library curation.} We remove task-specific usage guidance or geometric verification guidance, or use append-only storage.}
\label{tab:ablation-results}
\end{table}

\begin{figure*}[!t]
    \centering
    \includegraphics[width=0.9\textwidth]{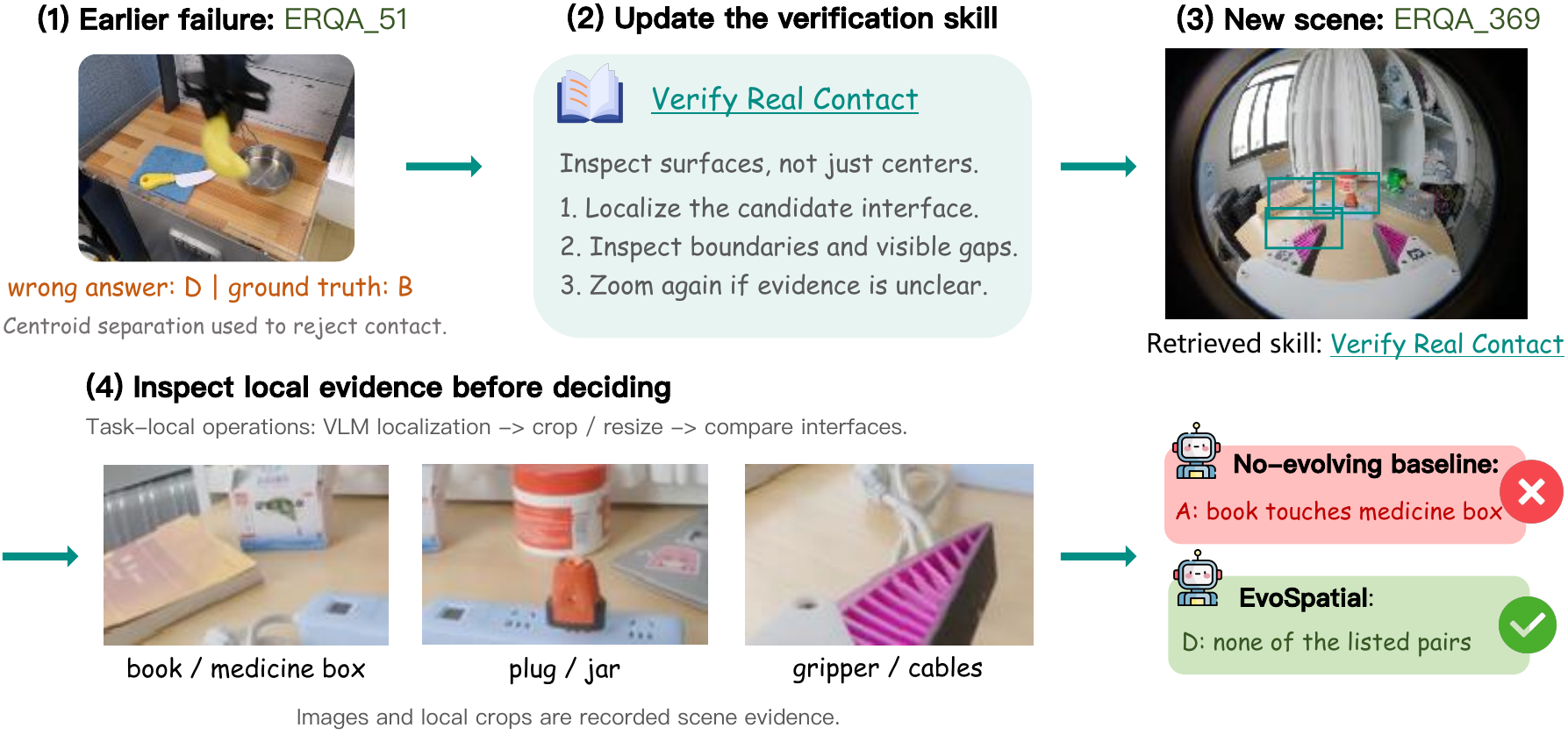}
    \caption{\textbf{From failure reflection to skill reuse.} A revised contact-verification skill directs the agent to inspect local surfaces before answering in a subsequent scene.}
    \label{fig:evolution-case-study}
\end{figure*}

\subsection{Cross-benchmark Transfer}
\label{sec:cross-benchmark-transfer}

\begin{figure}[t]
    \centering
    \includegraphics[width=0.85\columnwidth]{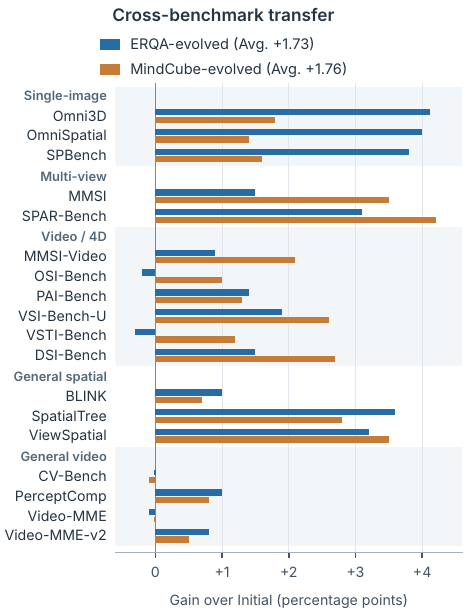}
    \caption{\textbf{Cross-benchmark transfer from ERQA and MindCube.}
    Score gains over the initial library on 18 common target benchmarks, excluding both sources. Libraries remain fixed during target evaluation.}
    \label{fig:erqa-transfer}
\end{figure}

We evolve separate libraries on ERQA and MindCube from the same initial library using Gemma4-31B, then freeze them for evaluation on 18 common targets excluding both sources. Relative to the fixed \emph{Initial} library, \emph{ERQA-evolved} and \emph{MindCube-evolved} improve the equally weighted target average by 1.73 and 1.76 points (Figure~\ref{fig:erqa-transfer}).

Transfer profiles differ: ERQA yields larger single-image gains (3.97 versus 1.60), while MindCube yields larger multi-view (3.85 versus 2.30) and video spatial/4D gains (1.82 versus 0.87). Manual inspection suggests ERQA's image-level procedures are seldom invoked in some video tasks. MindCube instead improves all six video spatial/4D targets, including those with small ERQA-transfer declines. Both sources yield smaller gains on general video understanding. Transfer therefore extends beyond ERQA but depends on compatibility with target requirements.

\subsection{Learning from Skill-Guided Trajectories}
\label{sec:skill-distillation}

To test the training value of evolved skills, Gemma4-31B generates Structured Tool-Call trajectories on SenseNova-SI-800K~\cite{cai2026sensenovasi} with or without fixed ERQA-evolved skills. Both conditions share tools and execution budgets, with evolution and helper synthesis disabled. Identical correctness and execution checks retain the same 200K questions passing under both conditions. Their trajectories train \emph{Unguided Trace SFT} and \emph{Skill-Guided Trace SFT} from Gemma4-31B with matched supervised token budgets and optimization settings. Skill documents are omitted from training inputs; only assistant actions and answers receive supervision. Reference answers serve only post-generation filtering.

Evaluation uses Structured Tool-Call without external skills or updates on nine benchmarks excluding ERQA; training excludes shared evaluation images and scenes. Table~\ref{tab:skill-distillation} includes the original \emph{Base Model}, aggregated from the main-table baseline. Skill-Guided Trace SFT improves over Unguided Trace SFT by $1.2$--$1.4$ points in every task group, demonstrating the training value of skills under matched questions and budgets.

\begin{table}[!ht]
\centering
\footnotesize
\setlength{\tabcolsep}{2.5pt}
\renewcommand{\arraystretch}{1.12}
\begin{adjustbox}{max width=\columnwidth}
\begin{tabular}{@{}>{\raggedright\arraybackslash}m{0.44\columnwidth}cccc@{}}
\toprule
Training condition & \makecell{Single-\\image} & \makecell{Multi-\\view} & General & \makecell{Avg.\\(9)} \\
\midrule
Base Model & 58.6 & 54.7 & 62.4 & 58.6 \\
\rowcolor{black!6}
Unguided Trace SFT & 59.6 & 55.9 & 63.2 & 59.6 \\
\rowcolor{white}
\textbf{Skill-Guided Trace SFT} & \textbf{60.8} & \textbf{57.3} & \textbf{64.4} & \textbf{60.9} \\
\bottomrule
\end{tabular}
\end{adjustbox}
\caption{\textbf{Learning from skill-guided trajectories.} The two SFT variants use trajectories generated with or without evolved skills for the same 200K questions.}
\label{tab:skill-distillation}
\end{table}

\subsection{Evaluation across Base Models}
\label{sec:backbone-evaluation}

Table~\ref{tab:backbone-results} evaluates \methodname{} with five Gemma and Qwen backbones on 10 benchmarks. Average scores range from 61.2 (Gemma4-26B-A4B) to 67.6 (Qwen3.8-27B).

\subsection{Token Efficiency}
\label{sec:token-efficiency}

Table~\ref{tab:token-efficiency} reports mean ERQA token costs for Gemma4-31B with SpatialClaw or \methodname{}, and Qwen3.8-27B with SpatialClaw, including \methodname{}'s post-answer evolution. With Gemma4-31B, \methodname{} improves accuracy by 2.2 points while reducing total tokens from 68.5k to 54.2k (20.9\%). Mean tool-call turns decrease from 5.09 to 4.32 (15.1\%), with reductions in both successful and failed episodes (Appendix~\ref{app:execution-steps}). Shorter interactions support reduced repeated-context input: input tokens fall from 65.8k to 50.8k despite the added cost of skill documents, selection, and evolution.

\begin{table}[!ht]
\centering
\footnotesize
\setlength{\tabcolsep}{3pt}
\renewcommand{\arraystretch}{1.12}
\begin{adjustbox}{max width=\columnwidth}
\begin{tabular}{@{}>{\raggedright\arraybackslash}m{0.37\columnwidth}cccc@{}}
\toprule
System & \makecell{Acc.\\(\%)} & \makecell{Input\\(k/question)} & \makecell{Output\\(k/question)} & \makecell{Avg. total\\(k/question)} \\
\midrule
Gemma4-31B\newline + SpatialClaw & 61.3 & 65.8 & 2.7 & 68.5 \\
\rowcolor{black!6}
Gemma4-31B\newline + \textbf{\methodname{}} & 63.5 & 50.8 & 3.4 & \textbf{54.2} \\
Qwen3.8-27B\newline + SpatialClaw & 63.0 & 145.9 & 17.6 & 163.5 \\
\bottomrule
\end{tabular}
\end{adjustbox}
\caption{\textbf{Token efficiency on ERQA.} Token counts are averaged per question.}
\label{tab:token-efficiency}
\end{table}

\section{Case Study}
\label{sec:case-study}

Figure~\ref{fig:evolution-case-study} shows reflection turning a faulty centroid-based contact judgment into a reusable instruction to inspect local surfaces and gaps. The revised skill guides evidence gathering in another scene, transferring a verification criterion rather than an answer.

\section{Conclusion}
\label{sec:conclusion}

We presented \methodname{}, which evolves linked spatial skills and tools from execution feedback, retaining geometric assumptions and verification steps alongside reusable computations. It exceeds SpatialClaw's reported average by 4.0 points across 20 benchmarks. Fixed-library evaluations, transfer from two sources, and ablations support both experience reuse and explicit verification guidance. Skill-guided trajectories also improve fine-tuning without external skills at evaluation. Checked spatial experience thus benefits both subsequent reasoning and model training.

\section*{Limitations}

\textbf{Runtime isolation.}
For rapid research iteration, the prototype executes agent-generated code without sandbox isolation. This code inherits the runtime's permissions and could damage the host or its data; we observed no such incidents. A compatible initial isolation scheme is to host the task-local Python kernel in a dedicated, non-privileged Docker container or Kubernetes Pod with restricted filesystem mounts, network access, and resource budgets. This preserves state across tool-call turns while limiting access to the host. We have not evaluated this deployment configuration or its security guarantees.

\textbf{Cross-benchmark transfer.}
Our cross-benchmark transfer analysis uses ERQA and MindCube as evolution sources. Model API costs limit the exploration of additional sources and combinations, leaving transfer across a broader range of source domains untested.

\section*{Ethics Statement}

\textbf{Data and benchmarks.}
We use publicly available benchmarks and collect no new human-subject data. We follow the source datasets' licenses and terms and introduce no personally identifiable or sensitive information.

\textbf{Code execution.}
The agent executes LLM-generated code to inspect visual and geometric evidence. As discussed in Limitations, execution is not sandboxed, so generated code could perform unintended operations with the runtime's permissions. We observed no harmful behavior during the study and recommend sandboxing before deployment.

\bibliography{custom}

\appendix

\section{Implementation of Experience Reuse}
\label{app:implementation}

We describe the retrieval, feedback, and curation settings used to manage the skill--tool library. Appendix~\ref{app:main-evolution-protocol} specifies the online evaluation protocol.

\paragraph{Retrieval.}
The resident layer contains at most two active, trusted skills explicitly marked as resident; it may be empty. On-demand retrieval combines BM25 with sentence embeddings from \texttt{BAAI/bge-small-en-v1.5}, retaining up to twelve candidates from each channel. We combine their ranks using reciprocal-rank fusion:
\begin{equation}
    R_{\mathrm{hyb}}(s,x)=
    \sum_{c\in\{\mathrm{sparse},\mathrm{dense}\}}
    \frac{\mathbf{1}[s\in C_c]}{60+\operatorname{rank}_c(s)},
\end{equation}
where $C_c$ is the candidate set from channel $c$, and a candidate absent from that channel contributes zero. The final ranking score is
\begin{equation}
    \begin{split}
    R(s,x) ={}& R_{\mathrm{hyb}}(s,x)+\lambda_a A(s,x)\\
              &+\lambda_r(2\rho_s-1)-p_{\mathrm{arch}}(s),
    \end{split}
    \label{eq:retrieval}
\end{equation}
where $A$ measures applicability, $\rho_s$ summarizes observed reliability, and $p_{\mathrm{arch}}$ penalizes archived entries. We set $\lambda_a=0.05$, $\lambda_r=0.02$, and the archival penalty to $0.01$. Reliability uses the smoothed estimate $\rho_s=(n_s^++1)/(n_s^++n_s^-+2)$, where $n_s^+$ and $n_s^-$ count supported and contradicted applications, respectively. Inconclusive applications increment neither count. We use statistics for matching task intents when available and global statistics otherwise. Applicability combines compatibility of operations, modalities, and capabilities. Incompatible inputs can exclude a candidate, while missing tools or capabilities can reduce its score. If dense retrieval is unavailable, the system uses sparse retrieval alone and records this fallback.

\paragraph{Execution interface.}
The planner receives the question, input metadata, tool descriptions, and selected skills. The execution agent reads each code cell's output before generating the next cell. Outputs include printed measurements, variable summaries, errors, and images registered through \texttt{show()}. Execution ends when the agent submits an answer through \texttt{ReturnAnswer()} or reaches the step limit.

\paragraph{Selection and synthesis.}
The selector receives at most eight candidates and chooses at most three on-demand skills. For each skill, it generates task-specific usage guidance covering relevant operations, unmet requirements, evidence to verify, and how the results should inform the answer. If the guidance is invalid or confidence in applicability is insufficient, the retrieved skill is withheld; resident skills remain available. By default, synthesis allows at most one pre-task proposal and one repair attempt per task. A repeated matching execution error triggers repair on its second occurrence. Admission checks inspect the Python abstract syntax tree, enforce import and operation restrictions, and compile and load the function before LLM curation. Execution checks on representative inputs verify the documented output structure and applicable computational properties. Repaired or merged helpers are checked against retained examples before replacing an existing implementation. Each generated helper returns a structured result containing success status, a value, a failure reason, and metadata.

\paragraph{Usage assessment.}
We assess skill application and helper utility from observed trajectories, without counterfactual rollouts. Each assessment records the skill or helper, its intended operation, the relevant observation or check, and a supported, contradicted, or inconclusive verdict. Positive evidence requires a checked result, such as a repaired input mismatch or a visually verified segmentation fallback. Errors and violated preconditions count as negative evidence only for the implicated skill or helper. Exposure alone, error-free execution, and confidence in the final answer do not establish utility. These assessments guide library management but do not establish causal benefit.

\paragraph{Curation and lifecycle.}
Skill updates make local changes supported by approved candidates, and a deterministic merge preserves unspecified fields. Usage statistics come from execution records and are maintained separately from curator-generated content. The active library holds at most 64 skills and 16 generated helpers. New entries have probationary status, and at most two probationary generated helpers are exposed at a time. By default, a skill can become trusted after two supported applications on distinct tasks. Promotion of a generated helper requires two evidence-backed helpful-use judgments, each with confidence of at least $0.70$. Confidence without an identified supporting check is insufficient for promotion. Contradictory evidence can trigger revision or quarantine; inconclusive judgments affect neither promotion nor demotion. Archiving preserves content, and archived skills remain eligible for retrieval with a score penalty. Selection, reflection, curation, and utility assessment use separate sessions of the configured backbone; they do not use independently trained evaluators.

\paragraph{Feedback protocol.}
During online adaptation, trajectory assessment and library updates finish before the next task begins. Reflection receives execution records, observed repairs, and visual or geometric checks. In held-out transfer evaluations, the library is built on a disjoint adaptation stream and remains fixed during testing. Token measurements include task-solving, verification, and evolution requests. The library distinguishes initial handwritten helpers from generated helpers.

\section{Evolution Protocol for the Main Results}
\label{app:main-evolution-protocol}

The main results in Table~\ref{tab:spatial-main-results} evaluate \methodname{} during online evolution under the following protocol.

\paragraph{Initialization and isolation.}
Each benchmark starts from a fresh copy of the initial skill--tool library. No evolved skills, generated helpers, or accumulated library records are shared across the 20 benchmarks. Each of the three runs also starts from the initial library, without state carried over from another run.

\paragraph{Sequential feedback and updates.}
The agent processes questions sequentially within each benchmark. After submitting an answer, it uses execution records and evidence checks to reflect and update the library before starting the next question. Only the offline scorer accesses benchmark labels; neither the agent nor the library-management modules receive them. Updated entries remain available to later questions in the same benchmark and run. Each score uses the answer submitted before the corresponding post-answer update.

\paragraph{Task order and reporting.}
The three runs use different question orders, forming separate evolution streams over the same benchmark examples. We report the mean and sample standard deviation of the three run-level scores. Category and overall averages are computed within each run before aggregation. These results measure performance during online evolution; Sections~\ref{sec:held-out-generalization} and~\ref{sec:cross-benchmark-transfer} separately evaluate reuse with a fixed library.

\section{Tool-Call Turns on ERQA}
\label{app:execution-steps}

We compare mean tool-call turns on ERQA for SpatialClaw and \methodname{} with Gemma4-31B and Qwen3.8-27B (Table~\ref{tab:execution-steps}). One turn is one executed agent code cell, including failed attempts and answer submission; multiple tool calls within a cell count as one turn. We average over all episodes and separately over those with correct and incorrect final answers, including unanswered episodes as failures. Selection, reflection, and library-update calls are excluded from turn counts but included in token costs.

\begin{table}[!ht]
\centering
\footnotesize
\setlength{\tabcolsep}{3pt}
\renewcommand{\arraystretch}{1.12}
\begin{adjustbox}{max width=\columnwidth}
\begin{tabular}{@{}>{\raggedright\arraybackslash}m{0.43\columnwidth}ccc@{}}
\toprule
System & \makecell{All\\episodes} & \makecell{Successful\\episodes} & \makecell{Failed\\episodes} \\
\midrule
Gemma4-31B\newline + SpatialClaw & 5.09 & 4.59 & 5.80 \\
\rowcolor{black!6}
Gemma4-31B\newline + \textbf{\methodname{}} & 4.32 & 4.24 & 4.46 \\
\rowcolor{white}
Qwen3.8-27B\newline + SpatialClaw & 10.30 & 7.89 & 14.98 \\
\rowcolor{black!6}
Qwen3.8-27B\newline + \textbf{\methodname{}} & 8.15 & 6.45 & 11.24 \\
\bottomrule
\end{tabular}
\end{adjustbox}
\caption{\textbf{Mean tool-call turns on ERQA.}}
\label{tab:execution-steps}
\end{table}

\methodname{} reduces mean turns from 5.09 to 4.32 with Gemma4-31B (15.1\%) and from 10.30 to 8.15 with Qwen3.8-27B (20.9\%). Reductions occur in both successful and failed episodes. They are larger among failures: 23.1\% for Gemma4-31B and 25.0\% for Qwen3.8-27B, compared with 7.6\% and 18.3\% among successes. Shorter interactions occur in both outcome groups, with the largest reductions on episodes that ultimately fail. This pattern supports the interpretation that reusable procedures help limit prolonged, unproductive interaction and reduce repeated context input.

\section{LLM Judge Reliability}
\label{app:judge-reliability}

Table~\ref{tab:judge-reliability} reports the Gemma4-31B judge's agreement with human judgments, acceptance of unsupported applications, and consistency across repeated judgments, separately for skills and generated helpers.

\begin{table}[!ht]
\centering
\footnotesize
\setlength{\tabcolsep}{3pt}
\renewcommand{\arraystretch}{1.12}
\begin{adjustbox}{max width=\columnwidth}
\begin{tabular}{@{}>{\raggedright\arraybackslash}m{0.28\columnwidth}ccc@{}}
\toprule
\makecell[l]{Assessed\\artifact} & \makecell{Human\\agreement $\uparrow$} & \makecell{Unsupported\\acceptance $\downarrow$} & \makecell{Repeat\\consistency $\uparrow$} \\
\midrule
Skills & 84.0 & 9.0 & 90.0 \\
\rowcolor{black!6}
Generated helpers & 88.0 & 6.0 & 94.0 \\
\rowcolor{white}
Overall & 85.3 & 8.0 & 91.3 \\
\bottomrule
\end{tabular}
\end{adjustbox}
\caption{\textbf{LLM judge reliability.} All values are percentages.}
\label{tab:judge-reliability}
\end{table}

On the audited records, the judge agrees with human judgments on 85.3\% of cases and repeats its own verdicts with 91.3\% consistency. These rates support its use as a largely human-aligned and repeatable source of local feedback for library management.

\section{Representative Prompt Templates}
\label{app:prompts}

The following templates cover skill selection, tool synthesis, reflection, and library curation. For readability, we omit lengthy JSON schemas and task-dependent tool documentation. Angle brackets mark content supplied at runtime, and terminology follows the main text.

\begin{promptbox}[title={Skill selection}]
\small
You are the Skill Selector for a spatial reasoning agent. The retriever favors recall, so some candidates may be distractors. Decide whether skills are needed, and select the smallest sufficient set of complementary skills. Check the input modality, preconditions, contraindications, required evidence, available tools, and requirements that each skill does not cover. Do not select a skill by topical similarity alone; selecting none is valid.

For each selected skill, return the operations for which it can be used, the requirements it does not cover, and the evidence needed before its advice can affect the answer.

\emph{Input:} \texttt{<question>}, \texttt{<task signature>}, \texttt{<resident skills>}, and \texttt{<retrieved candidates>}. \emph{Output:} selected skill IDs, task-specific usage guidance, rejected candidates, and a brief rationale.
\end{promptbox}

\begin{promptbox}[title={Skills and task-specific usage guidance supplied to the execution agent}]
\small
These skills are reusable strategies distilled from checked operations and observed repairs. Treat them as optional guidance whose applicability must be verified. First identify the task requirements independently. Use a skill only when its conditions match the task, and verify the visual and geometric evidence before allowing it to affect the answer. If the required evidence is missing or contradictory, solve the task using other evidence.

\emph{For each selected skill:} \texttt{<when to use>}; \texttt{<when not to use>}; \texttt{<procedure>}; \texttt{<verification>}; \texttt{<failure modes>}; \texttt{<recommended tools>}; and \texttt{<task-specific usage guidance>}.
\end{promptbox}

\begin{promptbox}[title={Demand-driven tool synthesis}]
\small
Decide whether the task lacks a reusable Python helper. Reuse existing spatial tools rather than wrapping or duplicating them. Propose a small deterministic function only when it would simplify repeated coordinate parsing, validation, aggregation, or numerical computation. Give it a documented signature, an example call, and tests. The helper must have one function definition and no file, network, or process access or global state.

\emph{Input:} \texttt{<question>}, \texttt{<input metadata>}, \texttt{<relevant existing tools and skills>}, and, when available, \texttt{<execution failure>}. \emph{Output:} whether a helper is needed, the reason, and the proposed function and interface.
\end{promptbox}

\begin{promptbox}[title={Experience reflection}]
\small
Identify a reusable lesson supported by the execution record. For each relevant operation, distinguish supporting evidence, an observed contradiction, and an unresolved judgment. Link every proposed lesson to a concrete output, visual observation, or executed check.

For a repaired failure, explain what failed, which change resolved it, and what observation verifies the repair. For a supported procedure, retain its preconditions and verification steps. A tool returning without error does not establish that the final answer is correct. If evidence is insufficient, return no candidate. Exclude task-specific answers and conclusions.

\emph{Input:} \texttt{<question>}, \texttt{<visual observations>}, \texttt{<compact trajectory>}, and \texttt{<execution checks>}. \emph{Output:} a supported skill candidate and its evidence, or no candidate.

\end{promptbox}

\begin{promptbox}[title={Library curation}]
\small
Integrate pending skill candidates into the existing library by adding, updating, merging, archiving, or rejecting them. Group skills by the underlying spatial reasoning pattern rather than surface wording. Prefer concise, operational guidance that can transfer across tasks; retain applicable conditions, verification signals, and useful failure modes. Do not retain sample-specific answers or conclusions. Every accepted update must identify its supporting candidates and execution evidence, and existing skills should be preserved unless explicitly archived or deprecated.

\textbf{Generated helper candidates.} Accept a generated helper only when its code matches its description, its interface is reusable, its execution checks pass, and it does not duplicate a stronger tool. Recheck merged implementations against retained examples. Preserve existing trusted tools unless they are obsolete, duplicated, or harmful.

\emph{Input:} \texttt{<existing skills and tools>} and \texttt{<new candidates>}. \emph{Output:} library updates and decisions on pending candidates.
\end{promptbox}

\section{An Evolved Skill and Its Linked Tools}
\label{app:artifact-example}

The example below pairs an evolved spatial reasoning procedure with its verification steps and linked tools.

\begin{promptbox}[title={Metric Spatial Relationship Verification}]
\small
\textbf{When to use.} Verify both a metric distance and a directional relation between a robot gripper and a target in a 3D scene.

\textbf{Procedure.} Reconstruct the scene and segment the gripper and target. Identify their relevant 3D reference points and a consistent direction axis. Compare the Euclidean distance between the points with the stated distance, and use the sign of their displacement projected onto the axis to check the directional relation.

\textbf{Verification.} Confirm that the distance is within tolerance and that the directional projection has the expected sign. Inspect the masks and coordinate frame if either check disagrees with the visual evidence.

\textbf{Linked tools.}
\begin{compactitemize}
\item \texttt{tools.Reconstruct.Reconstruct}(frames) supplies depth and 3D geometry.
\item \texttt{tools.Evolved.}\allowbreak\texttt{segment\_with\_fallback}(image, prompt, \ldots) supplies object masks, retrying with box grounding when text segmentation fails.
\end{compactitemize}
\end{promptbox}

The library records these tool links separately, allowing the skill to recommend operations without embedding their implementations.

\section{Benchmark Scoring}
\label{app:benchmark-scoring}
Table~\ref{tab:spatial-benchmark-metrics} lists the metrics used for the benchmark suite adopted from SpatialClaw~\cite{cho2026spatialclaw}.
Categorical questions receive binary correctness scores. Numerical questions use mean relative accuracy (MRA) with the thresholds specified by the source benchmark. SPAR-Bench also uses view-change inference (VCI), which averages MRA over five movement axes. We average example scores within each benchmark and weight benchmarks equally when computing the overall score.

\begin{table}[t]
\centering
\small
\setlength{\tabcolsep}{3pt}
\begin{adjustbox}{max width=\columnwidth}
\begin{tabular}{@{}ll@{}}
\toprule
Benchmark & Metric \\
\midrule
ERQA~\cite{team2025gemini} & Acc \\
Omni3D~\cite{marsili2025visual} & Acc + MRA \\
OmniSpatial~\cite{jia2025omnispatial} & Acc \\
SPBench~\cite{li2025spatialladder} & Acc + MRA \\
MindCube~\cite{yin2025mindcube} & Acc \\
MMSI~\cite{yang2025mmsi} & Acc \\
SPAR-Bench~\cite{zhang2025from} & Acc + MRA + VCI \\
MMSI-Video~\cite{lin2025mmsivideo} & Acc \\
OSI-Bench~\cite{wu2025osibench} & Acc + MRA \\
PAI-Bench~\cite{zhou2025paibench} & Acc \\
VSI-Bench-U~\cite{brown2025shortcuts} & Acc + MRA \\
VSTI-Bench~\cite{fan2025vlm} & Acc + MRA \\
DSI-Bench~\cite{zhang2025dsibench} & Acc \\
BLINK~\cite{fu2024blink} & Acc \\
SpatialTree~\cite{xiao2025spatialtree} & Acc + MRA \\
ViewSpatial~\cite{li2025viewspatial} & Acc \\
CVBench~\cite{zhu2025cvbench} & Acc \\
PerceptionComp~\cite{li2026perceptioncomp} & Acc \\
Video-MME~\cite{fu2025video} & Acc \\
Video-MME-v2~\cite{videommev2_2026} & Acc \\
\bottomrule
\end{tabular}
\end{adjustbox}
\normalsize
\caption{\textbf{Benchmark scoring protocols.} Acc: categorical accuracy; MRA: mean relative accuracy; VCI: view-change inference.}
\label{tab:spatial-benchmark-metrics}
\end{table}

\end{document}